\documentclass[11pt,letterpaper]{article}
\usepackage{naaclhlt2016}
\usepackage{times}
\usepackage{url}
\usepackage{latexsym}
\usepackage{amsmath}
\usepackage{graphicx}
\usepackage{color}
\usepackage{booktabs}
\usepackage{subcaption}
\usepackage{xspace}
\usepackage{fix2col}

\newcommand{\scr}{\emph{StreetCrowd\xspace}}
\newcommand{\xhdr}[1]{{\noindent\bfseries #1}}
\newcommand{\Btask}{({++})}
\newcommand{\Ctask}{({+})}
\newcommand{\Wtask}{({-\,-})}
\newcommand{\score}{score\xspace}

\newcommand{\team}{team\xspace}
\newcommand{\teams}{teams\xspace}
\newcommand{\location}{location\xspace}

\newcommand{\puzzles}{puzzles\xspace}
\newcommand{\puzzle}{puzzle\xspace}

\newcommand{\sep}{\noalign{\vskip 0.8mm}}

\naaclfinalcopy

\title{Conversational Markers of Constructive Discussions}
\author{Vlad Niculae \and Cristian Danescu-Niculescu-Mizil\\
\small {\tt \{vlad|cristian\}@cs.cornell.edu}\\
        Cornell University}
\begin{document}
\maketitle
\begin{abstract}

Group discussions are essential for organizing every aspect of modern life, from faculty meetings to senate debates, from grant review panels to papal conclaves.
While costly in terms of time and organization effort, group discussions are commonly seen as a way of reaching better decisions
compared to solutions that do not require coordination between the
individuals (e.g.\ voting)---through discussion,
the sum becomes greater than the parts. 
However, this assumption is not irrefutable: anecdotal evidence of wasteful discussions abounds, and in our own experiments we find that over 30\% of discussions are
  unproductive.

We propose a framework for analyzing conversational dynamics in order to determine 
whether a given task-oriented discussion is worth having or not.
We exploit
conversational patterns reflecting the {\em flow of ideas} and the {\em balance} between the participants, as well as their linguistic choices.
We apply this framework to
conversations naturally occurring
in
an online collaborative 
 world exploration game developed and deployed to support this research.
Using this setting, we show that linguistic cues and conversational patterns extracted from the first 20 seconds of a team discussion are predictive of whether it
will be a wasteful or a productive one.

\end{abstract}

\section{Introduction}\label{sec:intro}

Working in {\teams} is a common
strategy for decision making and problem solving,
as building on effective
social interaction and on the abilities of each member can enable a {\team} to
outperform lone individuals.
Evidence shows that {\teams} often perform better than individuals \cite{williams1988group}
and even have high chances of reaching correct answers when all {\team} members
were previously wrong \cite{laughlin1980social}.
Furthermore, {\team} performance is not a factor of individual intelligence, but of
{\em collective intelligence} \cite{woolley2010evidence},
with interpersonal interactions and emotional intelligence playing
an important role \cite{jordan2002workgroup}.

Yet, as most people can attest from experience, {\team} interaction is not always smooth, and
poor
coordination 
can lead to 
unproductive
 meetings and wasted time. In fact, 
\newcite{romano2001meeting} report that one third of work-related meetings in the U.~S.\ are considered unproductive,  
while a 2005 Microsoft employee survey reports that 69\% of meetings
are ineffective.\footnote{\scriptsize\url{money.cnn.com/2005/03/16/technology/survey/}} As such, many grow cynical of meetings. 

Computational methods with the ability to reliably recognize 
unproductive discussions could have an important impact on our society.
Ideally, such a system could provide
 actionable information as a discussion progresses,
indicating whether it
 is likely to turn out to be productive, rather than a waste of time.
In this paper we focus on the conversational aspects of productive interactions and take the following steps:
\begin{itemize}
\item introduce a {\em constructiveness} framework that allows us to characterize
{\teams} where discussion enables better performance than the individuals could reach,
and, conversely, {\teams}
better off not having discussed at all (Section \ref{sec:constr});

\item create a setting that is conducive to decision-making discussions, where all steps of the
process (e.g., individual answers, intermediate guesses) are
observable
to researchers:
the {\scr} game (Sections \ref{sec:data}--\ref{sec:constrinstreet});

\item develop a novel framework for conversational analysis in small group discussions,
studying aspects such as the flow of ideas, conversational dynamics, and group balance (Sections~\ref{sec:feat}--\ref{sec:predict}).
\end{itemize}

We reveal differences in the collective decision
process characteristic of
productive and unproductive teams, and show that these differences are reflected in their conversational patterns.
For example, the language used when new ideas are introduced and adopted encodes important discriminative cues.
Measures of interactional balance and language matching \cite{niederhoffer2002linguistic,DanescuMark} also prove to be informative,
suggesting that more balanced discussions are most productive.
Our results underline the potential held by computational approaches to conversational dynamics. To encourage further work in this direction, we 
 render our
dataset
of task-oriented discussions 
and our feature-extraction code
publicly available.%
\footnote{\scriptsize\url{https://vene.ro/constructive/}}

\section{Related Work}
\label{sec:relwork}

Existing 
computational work on task-oriented group interaction is largely focused on how well 
the
{\team}
performs.
\newcite{coetzee2015structuring} deployed and studied the impact of a chat-based {\team}
interaction platform in massive open
online
courses, finding that teams reach more
correct answers than individuals, and that the experience is more enjoyable.
One often studied experimental setting is the HCRC Map Task Corpus \cite{anderson1991hcrc},
consisting of 128 conversations between pairs of people, where a designated one gives
directions to the other. This simplified setting avoids issues like role establishment and
leadership.
\newcite{reitter2007predicting} find that successful dialogs are characterized by
long-term adaptation and alignment of linguistic structures at syntactic,
lexical and character level. A notable feature of this work is
the
{\em success prediction}
task attempted using only the first 5 minutes of conversation.
Other attempts use authority level features inspired from negotiation theory,
experimental meta-features, task-specific features \cite{mayfield2011data}, and sociolinguistic spelling differences \cite{Mayfield:2012:CRD:2160881.2160893}.
Another research path 
uses
negotiation tasks from the Inspire dataset
\cite{kersten2003mining},
a collection of 1525 online bilateral negotiations where
roles are fixed (buyer and seller) and success is defined by the sale going through.
\newcite{sokolova2008telling} use a bag-of-words model and investigate the
importance of temporal aspects.
\newcite{sokolova2012much}
measure
informativeness, quantified by lexical sets of
degrees, scalars and comparatives.

Research on success in groups with more than two
members is less common. \newcite{Friedberg:SpokenLanguageTechnologyWorkshop:2012}
model the grades of 27 group assignments from a class using 
measures of average entrainment,
finding task-specific words to be a strong cue.
\newcite{jung2011engineering} shows how the
affective balance expressed
in
{\teams} correlates with performance on engineering tasks, in 30 teams
of up to 4 students.
In a related study the balance in the first 5 minutes of an interaction
is found predictive of performance \cite{Jung:2012:GHB:2207676.2208523}.
None of the research we are aware of controls for initial skill or potential
of the {\team} members.

In management science, network analysis reveals that certain subgraphs found
in long-term, structured {\teams} indicate better performance, as rated
by senior managers \cite{Cummings:SocialNetworks:2003}; controlled
experiments show that optimal structures depend on the complexity of the
task \cite{Guetzkow:ManagementScience:1955,bavelas1950communication}.
These studies, as well as much of the research on effective {\team}
crowdsourcing \cite[{\em inter alia}]{lasecki2012real,wang2011diversity},
do not focus on linguistic and conversational factors.

\section{Constructive Discussions}\label{sec:constr}

The first hurdle is to reliably quantify how productive group conversations are.  
In problem-solving, the ultimate goal is to find the correct answer, or, failing
that, to come as close
to it
as possible.  To quantify closeness to the correct answer,
a {\em \score} is often used, such that better
guesses
get higher scores;
for example, school grades.

\begin{figure}[t]
    \centering
    \includegraphics[width=0.48\textwidth]{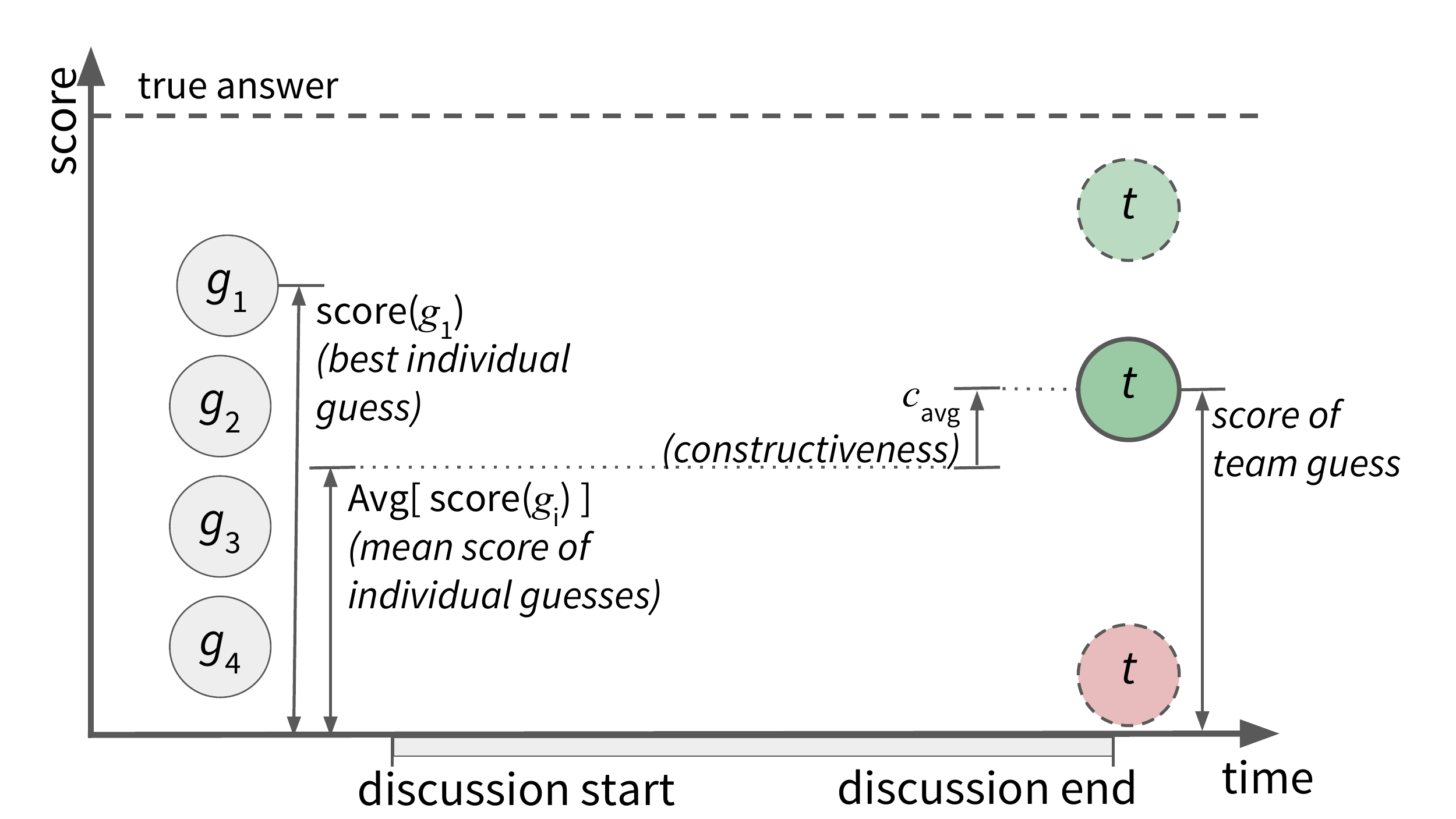}
    \caption{Intuitive sketch for constructiveness. The solid green circle corresponds a  team guess following a constructive discussion ($c_{\text{avg}}>0$), the dashed green circle corresponds to the scenario of a team that outperforms its best member ($c_{\text{best}}>0$), while the dashed red circle corresponds to a team that underperforms its worst member \mbox{($c_{\text{worst}}<0$)}.}
    \label{fig:constrsk}
\end{figure}

In contrast, our goal is to measure how productive a team's interaction is.
Scores are measures of correctness, so using them
as a proxy for interaction quality is not ideal: a team of straight {\em A}
students
can manage to
get an {\em A} on a project without
exchanging ideas,
while a group of {\em D} students getting a {\em B} is
more interesting. 
In the
latter case, the team's improved performance is likely to come from a good discussion and
an efficient exchange of complementary ideas---making the sum greater than the parts.
To capture this intuition we 
say a team discussion is {\em constructive} if it results in an
improvement over the potential of the individuals. 
We can 
then
quantify the degree of
constructiveness $c_{\text{avg}}$ 
as the improvement of the team score $t$ over the mean of the initial scores $g_i$ of the $N$ individuals in the team:
\[ c_{\text{avg}} = \operatorname{score}(t) - \frac{\sum_{i=1}^N\operatorname{score}(g_i)}{N}. \]
The higher $c_{\text{avg}}$ is, the more the team's answer, \textit{after discussion}, improves upon
the 
individuals'
 average performance
 \textit{before discussion};
zero constructiveness ($c_{\text{avg}}=0$) means the team performed no better than its members did
before discussing,
while negative constructiveness ($c_{\text{avg}}<0$) corresponds to
non-constructive
discussions.\footnote{From an operational perspective, a {\team} can choose, instead of
having a discussion, to aggregate individual guesses, e.g., by majority voting
or averaging.
Non-constructive discussions roughly correspond to cases where such an aggregate guess
would actually be better than what the \team discussion would accomplish.
}
Figure~\ref{fig:constrsk} sketches the idea visually: 
the dark green circle corresponds to the team's score after a constructive discussion ($c_{\text{avg}}>0$), being above the average individual score. 

Since individuals answers can sometimes vary widely,
we
 also
consider the extreme cases of teams that perform better than the best team member ($c_{\text{best}} > 0$) and worse than the worst 
member
 \mbox{($c_{\text{worst}} < 0$)}, where:
\begin{align*}
c_{\text{best}} & = \operatorname{score}(t) - \max_i \operatorname{score}(g_i) \\
c_{\text{worst}} & = \operatorname{score}(t) - \min_i \operatorname{score}(g_i).
\end{align*}
One way to think of the extreme cases is to imagine a team supervisor that collects the individual answers and aggregates them, without any external information.  An oracle supervisor can do no better than choosing the best answer. The discussion and interaction of teams where $c_{\text{best}} > 0$ leads to a better answer than such an oracle could achieve. 
(One such scenario is illustrated by the dashed
light green circle in Figure~\ref{fig:constrsk}.)
Similarly, teams where $c_{\text{worst}} < 0$ waste their time completely, as simply picking one of their members' answers at random is guaranteed to do better. (The dashed red circle in Figure~\ref{fig:constrsk} illustrates this scenario.)

The most important aspect of the constructiveness framework, in contrast to traditional measures of
correctness or success, is that all constructiveness
measures are designed to control for initial performance
or potential
of the team members, in order to focus on the
effect of the
discussion.\footnote{
Due to its relative nature, constructiveness also accounts for 
 variation in task difficulty in most scenarios. For example, in terms of $c_{\text{worst}}$,  when a team cannot even match its worst performing member, this is a sign of 
  poor team interaction even if the task is particularly challenging.}

In settings of importance, the true answer is not known 
a priori, and this constructiveness
cannot
be calculated directly.
 We therefore seek out 
 to model
  constructiveness
using
observable
 conversational
and linguistic correlates 
(Sections~\ref{sec:feat}--\ref{sec:predict}).
To develop such
  a model, 
   we design
   a large-scale experimental setting where the true answer is available to researchers,
but
unknown by the players
(Section~\ref{sec:data}).

\section{Experimental setting}\label{sec:data}
\subsection{StreetCrowd}
In order to study the constructiveness of task-oriented group discussion, we need a  setting that is conducive to decision-making discussions, where all steps of the
process (individual answers, intermediate guesses, group discussions and decisions) are
observable.  
Furthermore, to study at
scale, we
need to find a class of complex tasks with known solutions that can be automatically generated, but that
cannot
be easily solved by simply querying search engines.

With these constraints
in mind, we built {\scr}, an online multi-player world exploration game.\footnote{%
{\scriptsize \url{http://streetcrowd.us/start}}

(the experiment was approved by the IRB).
}
{\scr} is played in teams of at least two players
and is built around a geographic puzzle: determining your location
based on first-person images from the ground level.\footnote{
We
embed
Google Street View data.} Each \location generates a new puzzle.

\xhdr{Solo phase.} Each player has 3 minutes to navigate the surroundings, explore, and
try to find clues.  This happens independently and without communicating.  At the end, the player
is  
asked to make a guess by placing a marker on the world's map, and is prompted for an explanation and for a confidence level.
The answer is not yet revealed.

\xhdr{Team phase.} The team must then decide on a single, common 
guess.
  To accomplish
this, 
all
teammates 
are placed
in a chatroom and 
are provided
 with a map and a shared marker.
Any player can move the marker
at any point during the discussion.
 The game ends when all players agree on the answer, or
when the time limit is reached. 
 An example discussion is given in Figure \ref{fig:example}.
Guesses are scored according to their distance to the true {\location} using the
spherical
law of cosines:
$$ \operatorname{score}(\text{guess}, \text{true}) = -R d(\text{guess}, \text{true}) $$
\noindent where $d$ is the arc distance on a sphere, and $R$ denotes
the radius of the earth, assumed spherical. 
The score is given by the negative distance in kilometers, such that higher means better.
To motivate players and emphasize collaboration, the main {\scr} page displays a leaderboard 
consisting of the best team players.

The key aspects of the {\scr} design are:
\begin{itemize}
\item The 
\puzzles
 are complex and can be generated automatically in large numbers;
\item The true answers are known
to researchers,
but hard to obtain without solving the {\puzzle},
allowing for objective evaluation of both individual and group performance;

\item Scoring is continuous rather than discrete, allowing us to quantify
degrees
of improvement %
and capture incremental effects;
\item Each
teammate
has a different {\em solo phase} experience
and background knowledge, making it possible for the group discussion to shed light on new ideas;
\item The \puzzles are engaging and naturally conducive to collaboration, avoiding the use of monetary incentives
that can bias behavior.
\end{itemize}

%
%
%
%
%
%
%
%
%
%

%  %
%
\subsection{Preprocessing}
In the first
8 months,
over 1400 distinct players participated in over 2800 {\scr} games.
We tokenize and part-of-speech tag the conversations.\footnote{
We use the {\em TweetNLP} toolkit \cite{tweetnlp}
with
a tag set developed for Twitter data.  Manual examination reveals this
approach to be well suited for online chat data.
}
Before analysis,
due to the public nature of the game, we perform several filtering and quality check steps.

\xhdr{Discarding trivial games.}  We
remove all games that
the developers took part in.
We filter games where the team fails to provide a guess,
where fewer than two team members
engage in the team chat, 
and
\puzzles
 with insufficient samples.

\xhdr{Preventing and detecting cheating.} The {\scr} tutorial asks players to avoid using external resources
to look up clues and get an unfair advantage. 
To prevent cheating,
we detect and block
chat messages that link to websites,
and
we
employ
cookies and user accounts to prevent
people from playing
the same 
\puzzle
 multiple times.
To identify games that slip through this net,
we flag cases where the team, or any individual player, guesses within 10 km of the correct answer,
and leaves
the window while playing.
We further remove
a small set of
games
where the players confess to cheating in the
chat.

\begin{figure}[t]
    \centering
    \includegraphics[width=0.48\textwidth]{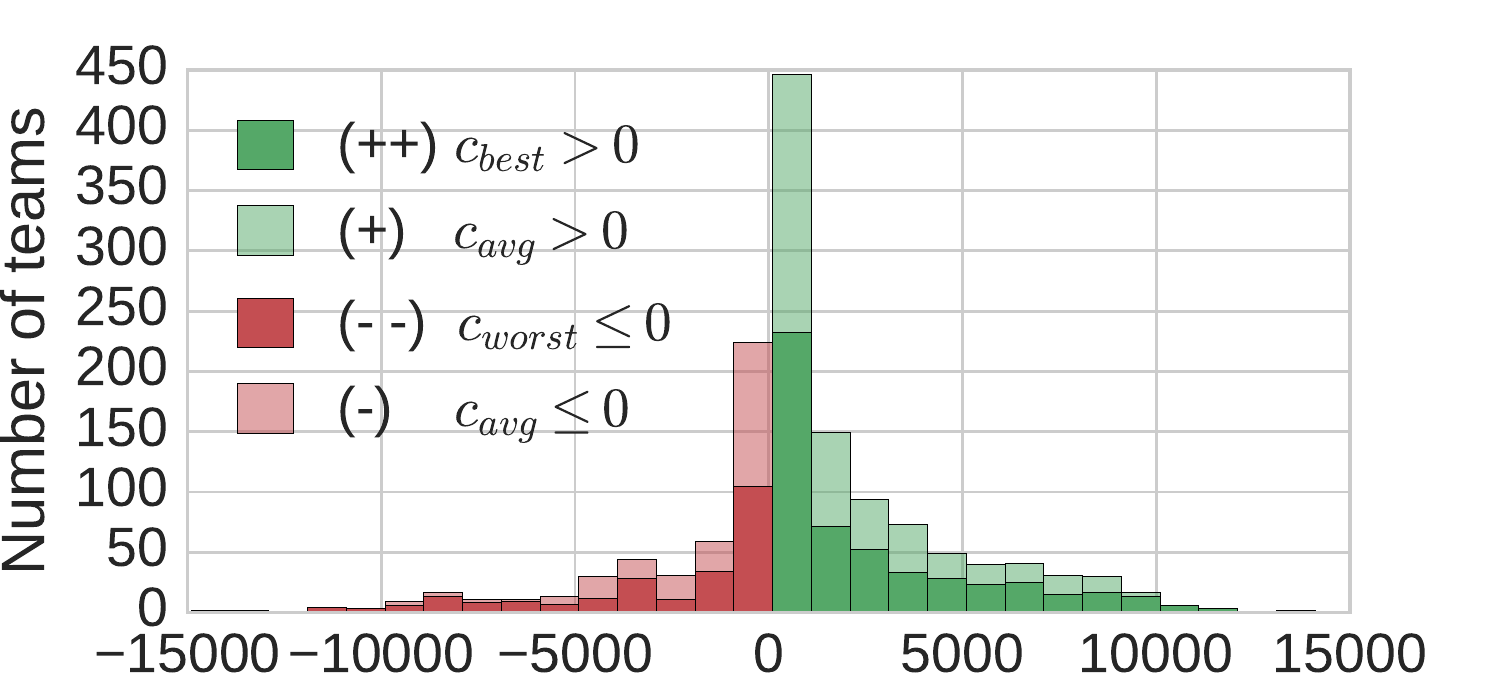}
    \caption{Distribution of {\team} constructiveness.}
    \label{fig:constr}
\end{figure}

\begin{figure}[t]
    \centering
	\captionsetup[subfigure]{justification=centering}
    \begin{subfigure}[t]{0.45\textwidth}
    \includegraphics[width=\textwidth]{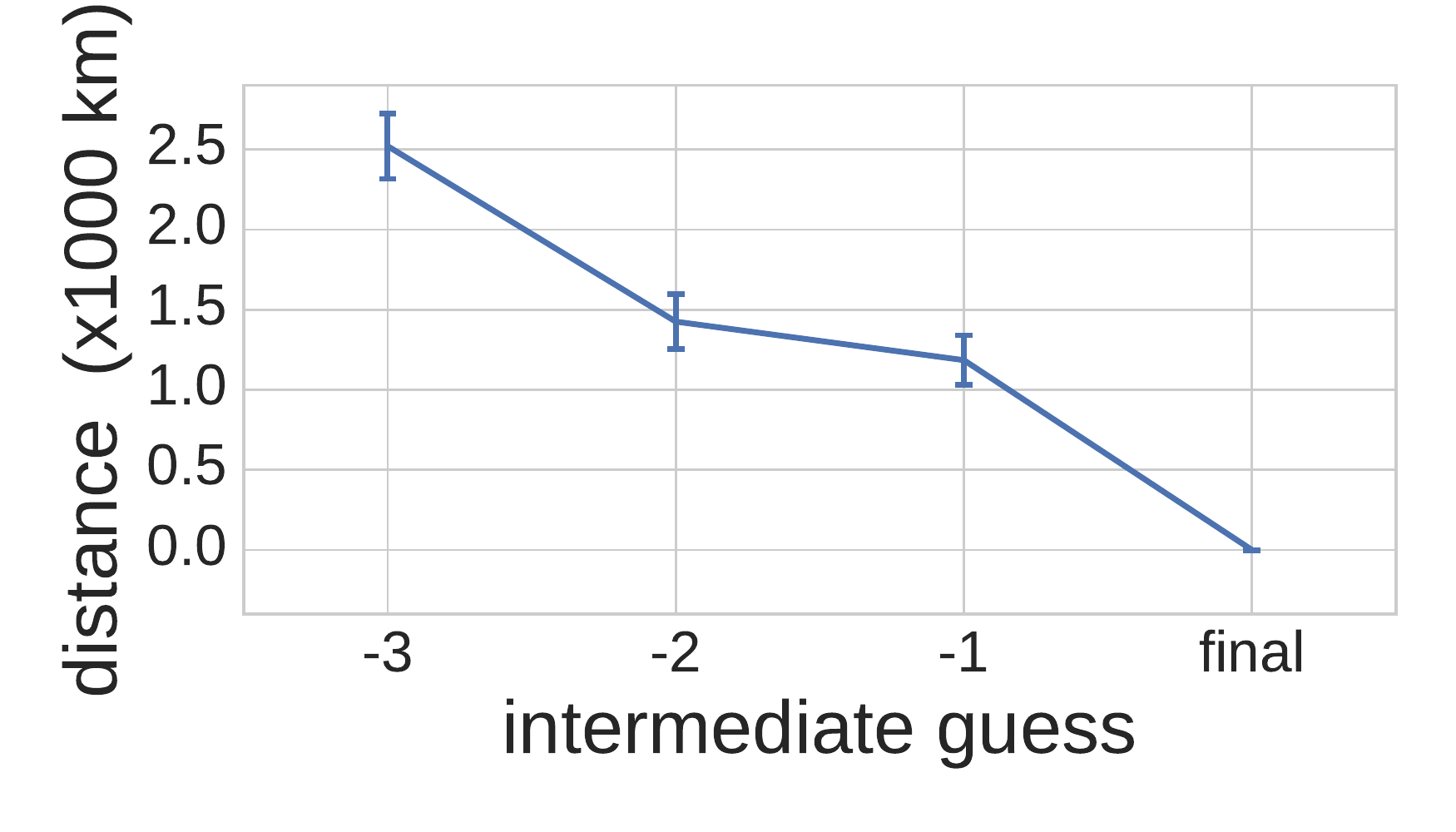}
    \caption{Distance between 
the last three
    intermediate guesses and the final guess.}
    \label{fig:marker_convergence}
    \end{subfigure}
    \begin{subfigure}[t]{0.45\textwidth}\
    \includegraphics[width=\textwidth]{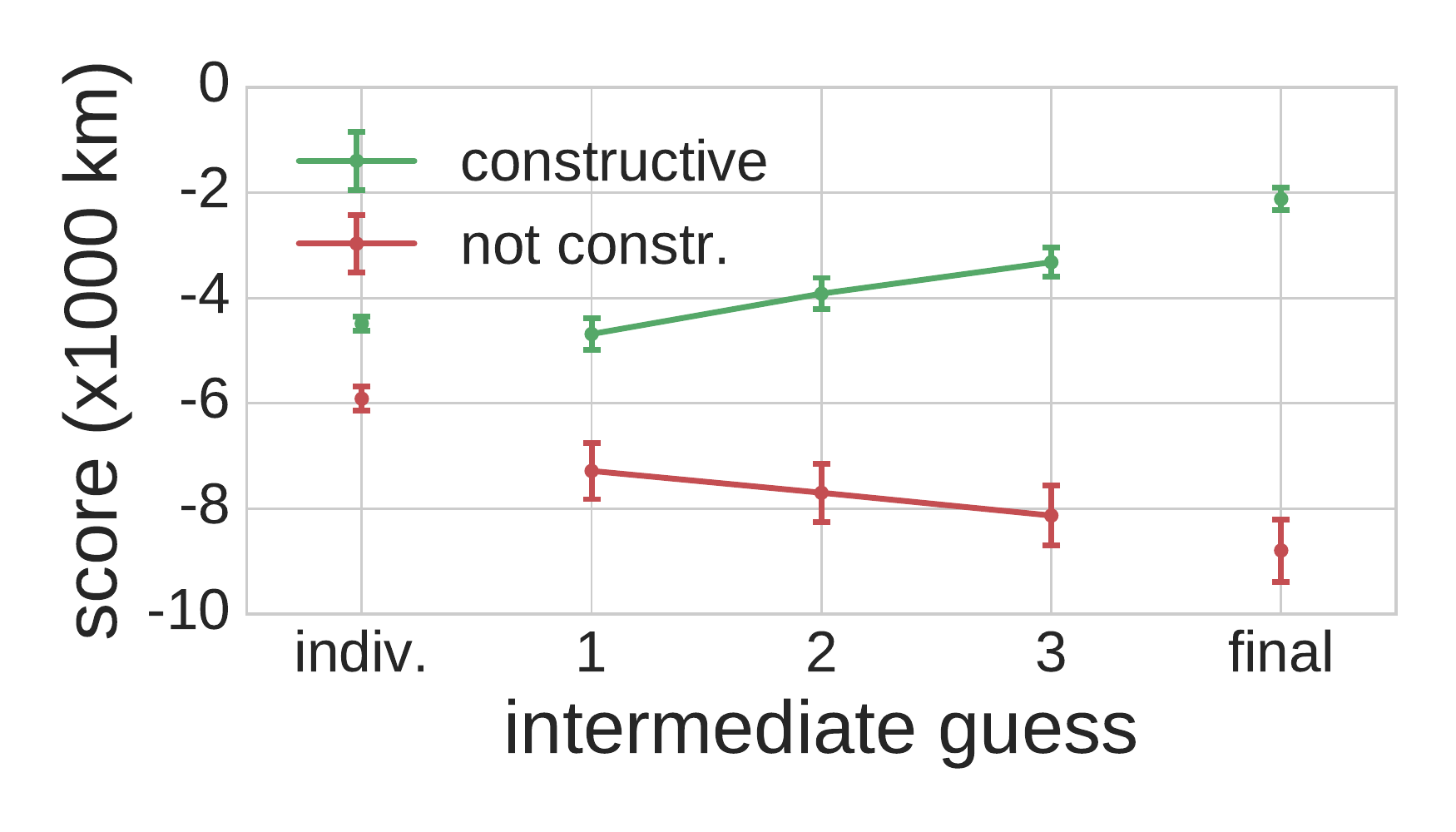}
    \caption{Score of the first three intermediate guesses; the mean score of the initial individual guesses and the score of the final team guess are shown for reference.}
    \label{fig:marker_score}
    \end{subfigure}
    \caption{Intermediate guesses offer a glimpse at the decision process:
    (a) guesses converge 
    rather than zig-zag  
    (b) in constructive games, guesses get incrementally better 
    than the mean individual score, 
    while in non-constructive games, they get worse.
    (Games with $\geq 3$ intermediate guesses.)
    }
\end{figure}
After filtering,
 our dataset
consists of 1450 games
on 70 different {\puzzles},
with an average of 3.9 games per
unique player, and
12.1 messages and 64.5 words in
an average
conversation.

\section{Constructiveness in StreetCrowd}\label{sec:constrinstreet}
We find that, indeed,
most of 
the
games are constructive. There are, however,
32\% %
non-constructive games ($c_{\text{avg}}<0$);
this reflects very closely the
survey by \newcite{romano2001meeting}.
Interestingly, 
in
36\%
of games, the team arrives at a better answer than any of the individual guesses  ($c_{\text{best}}>0$).  The
flip side is also remarkably common, with
17\%
of teams performing even worse than the worst individual  ($c_{\text{worst}}<0$).
The distribution of constructiveness is shown in Figure~\ref{fig:constr}: the fat tails indicate that
cases of large improvements and large deterioration are not uncommon.

\xhdr{Collective decision process.} Due to the full instrumentation of the game interface, we can investigate how constructiveness emerges out of the
{\team}'s interaction.
The {\team}'s intermediate guesses
during discussion confirm
that a meaningful 
process leads to the final {\team} decision:
guesses get closer and closer to the
final submitted 
guess
(Figure~\ref{fig:marker_convergence})%
; in other words, the {\team} {\em converges} to 
their
 final 
guess.

Notably, when considering how {\em correct} the intermediate guesses are, we notice an important difference between the way constructive and non-constructive {\teams} converge to their final guess 
(Figure \ref{fig:marker_score}).
  During their collaborative decision process,  constructive {\teams} make guesses that get closer and closer to the correct answer; in contrast, non-constructive {\teams} make guesses that
take them
  farther
  from the correct answer.
This observation has two important consequences. First, it shows that the two types of {\teams} behave differently
throughout,
suggesting
we could potentially detect
non-constructive discussions
early on, using interaction patterns.  Second, it emphasizes the potential practical value of such a task: 
 stopping a non-constructive {\team} early could lead to a better answer than
if they would carry on.

\begin{figure}[!ht]
    \centering
    \includegraphics[width=0.52\textwidth]{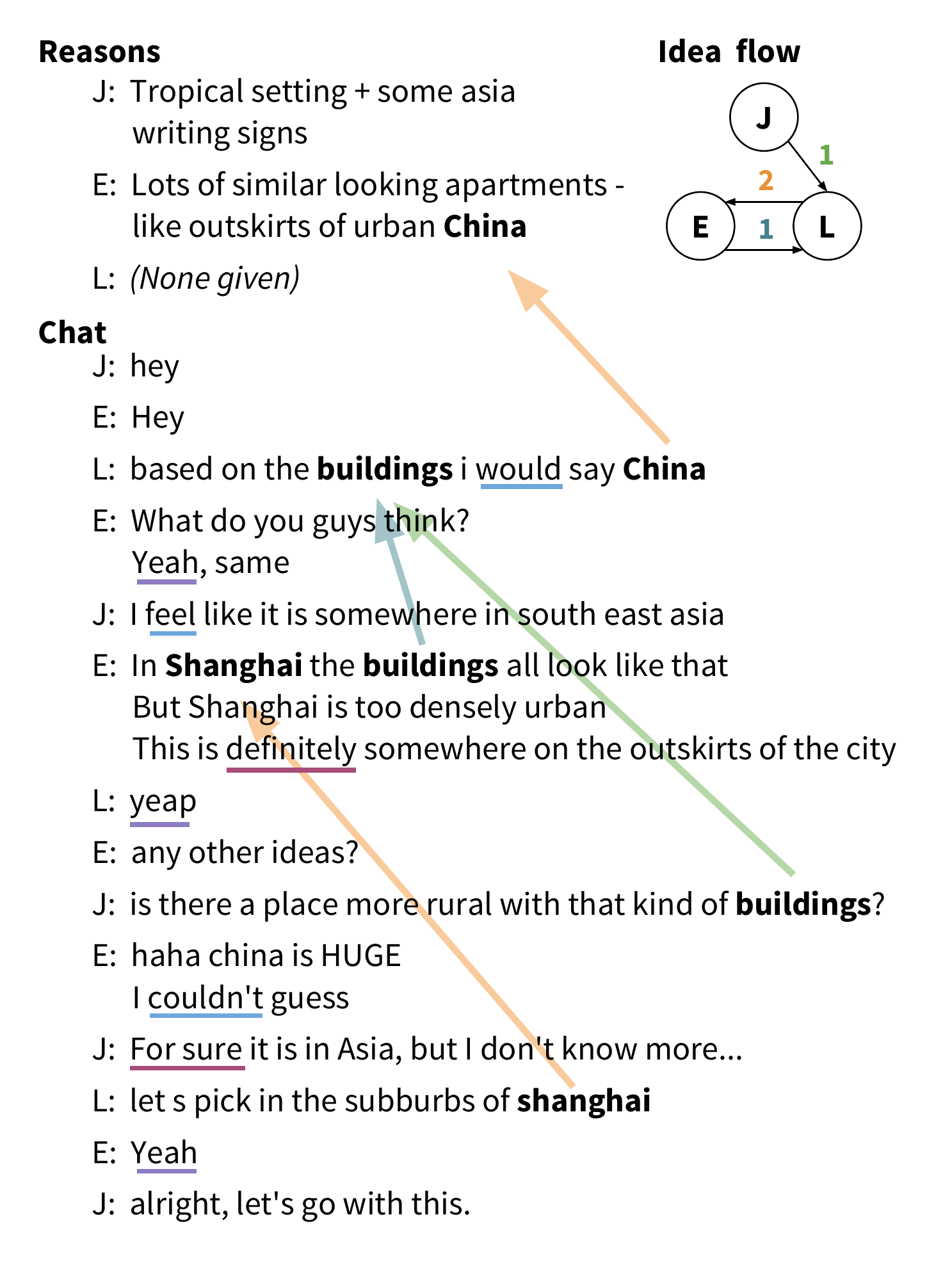}
    \caption{Example (constructive) conversation and the corresponding flow of ideas.
    Idea mentions are
    in bold,
    and relevant
    word classess
    are underlined.
    Arrow colors map to introducer--adopter pairs,
    matching the edges in the top-right graph.
    \label{fig:example}}
\end{figure}

\section{Conversation analysis}\label{sec:feat}

The process of \team convergence revealed in the previous section suggests 
a relation between
 the interaction leading to the final group decision and the relative quality of the outcome.
In this section, we develop a conversation analysis framework aimed at
characterizing
this relation.
This framework relies on conversational patterns and linguistic features, while steering away from lexicalized cues that might not generalize well beyond our experimental setting. 
To enable reproducibility, we make available the feature extraction code and the hand-crafted resources on which it relies.%
\footnote{\scriptsize\url{https://vene.ro/constructive/}}

\begin{figure*}[!ht]\centering
\captionsetup[subfigure]{justification=centering}
\begin{subfigure}[t]{0.195\textwidth}
\includegraphics[width=\textwidth]{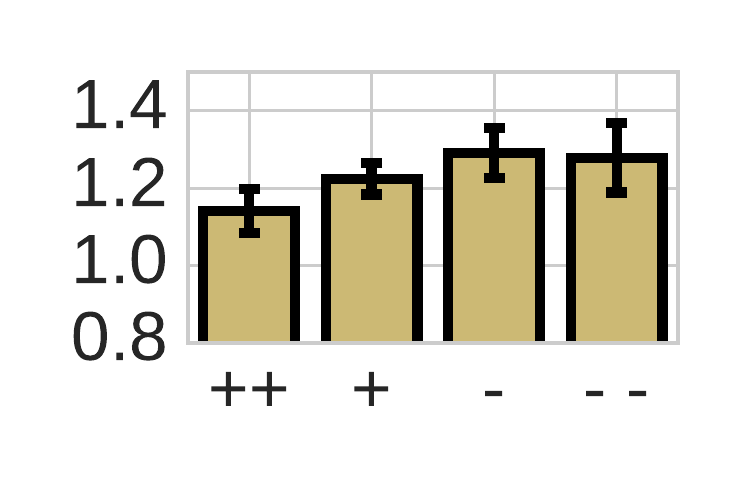}
\caption{Ideas by most\\prolific player\label{fig:maxideas}}
\end{subfigure}
\hfill
\begin{subfigure}[t]{0.195\textwidth}
\includegraphics[width=\textwidth]{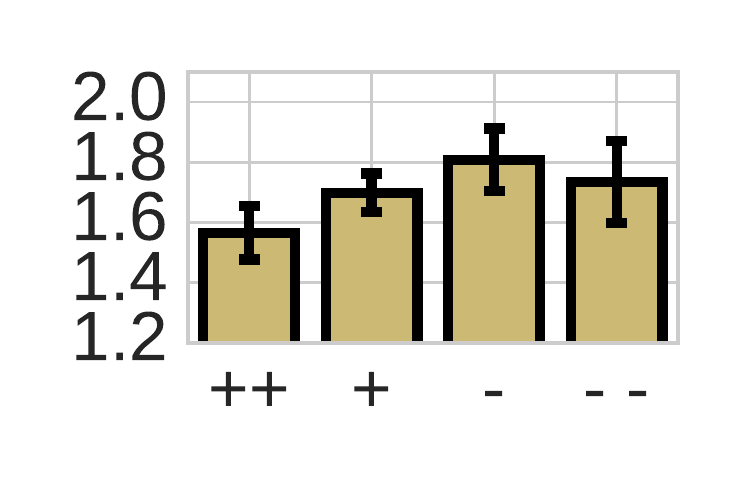}
\caption{Ideas adopted\label{fig:nideas}}
\end{subfigure}
\hfill
\begin{subfigure}[t]{0.195\textwidth}
\includegraphics[width=\textwidth]{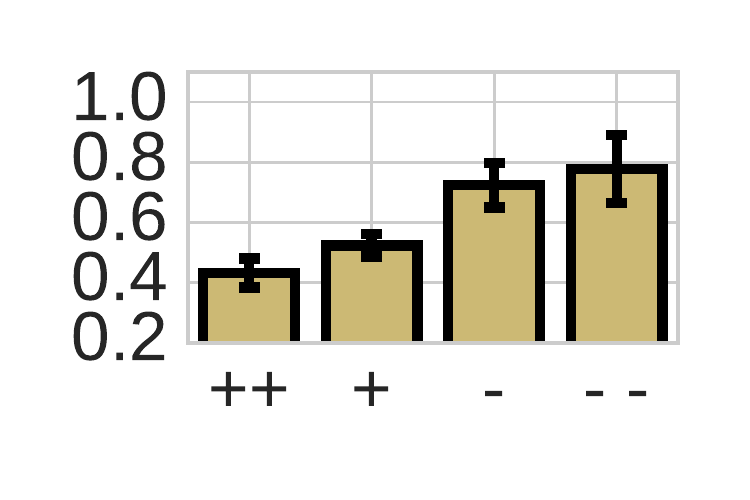}
\caption{Hedged idea\\introductions\label{fig:introhedge}}
\end{subfigure}
\hfill
\begin{subfigure}[t]{0.195\textwidth}
\includegraphics[width=\textwidth]{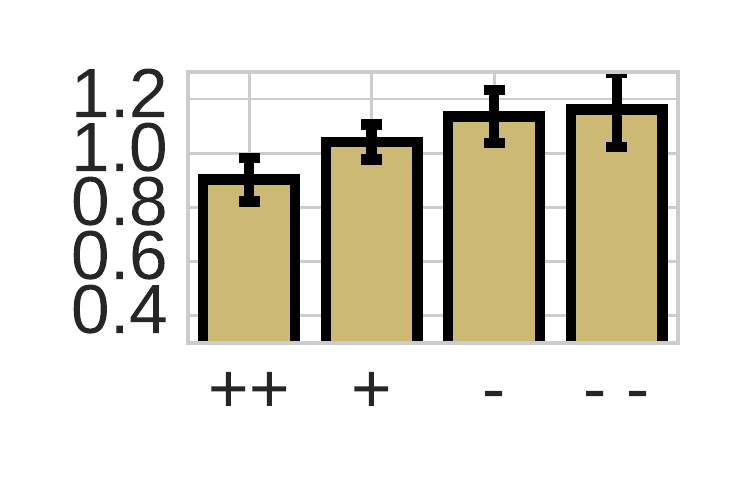}
\caption{Hedged idea\\adoptions\label{fig:adopthedge}}
\end{subfigure}
\hfill
\begin{subfigure}[t]{0.195\textwidth}
\includegraphics[width=\textwidth]{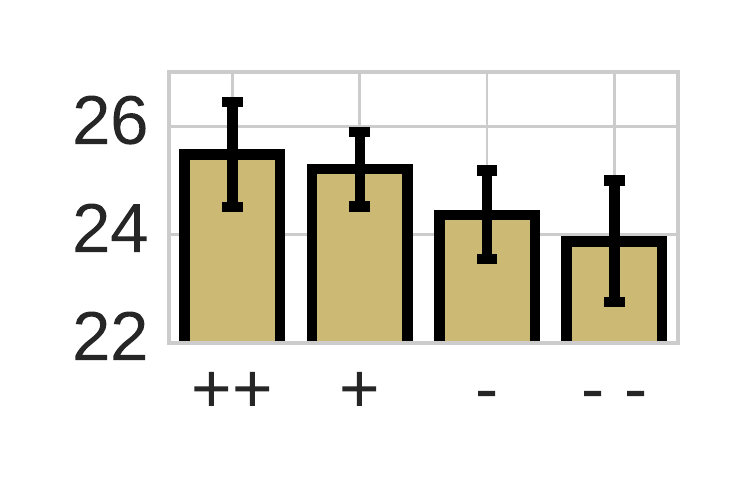}
\caption{Time between turns (seconds)}
\end{subfigure}

\begin{subfigure}[t]{0.195\textwidth}
\includegraphics[width=\textwidth]{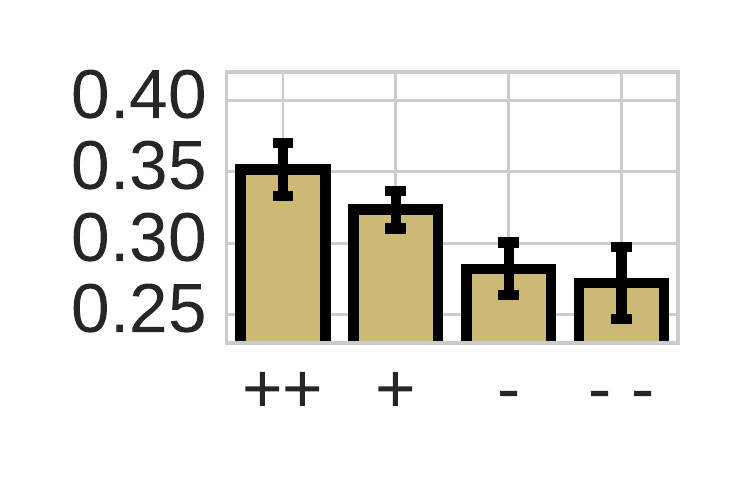}
\caption{Guessing entropy\label{fig:entmove}}
\end{subfigure}
\hfill
\begin{subfigure}[t]{0.195\textwidth}
\includegraphics[width=\textwidth]{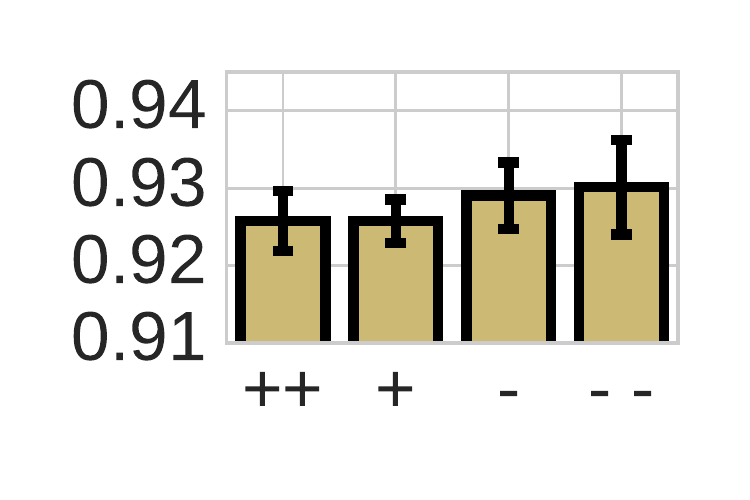}
\caption{Message entropy\label{fig:entmsg}}
\end{subfigure}
\hfill
\begin{subfigure}[t]{0.195\textwidth}
\includegraphics[width=\textwidth]{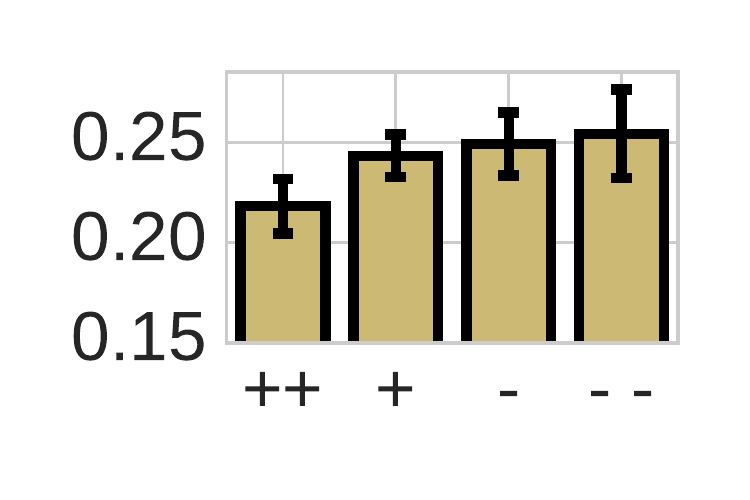}
\caption{Max-pair content word matching
\label{fig:maxaccom}}
\end{subfigure}
\hfill
\begin{subfigure}[t]{0.195\textwidth}
\includegraphics[width=\textwidth]{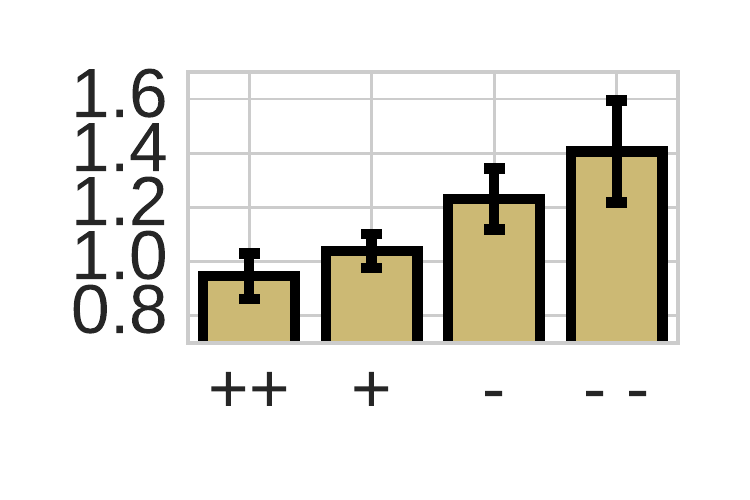}
\caption{%
Overall POS bigram matching\label{fig:meanaccom}}
\end{subfigure}
\hfill
\begin{subfigure}[t]{0.195\textwidth}
\includegraphics[width=\textwidth]{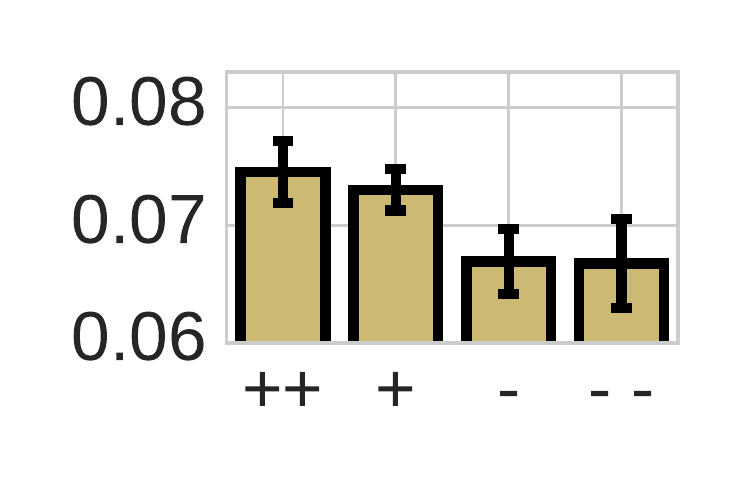}
\caption{Ratio of words related to geography\label{fig:geolex}}
\end{subfigure}

\caption{\label{fig:features}
 Averages for
 some of the predictive features,
based on idea flow (a-d), balance (e-i), and lexicons (j).
Error bars denote standard errors.
Legend: {\Btask}: teams that do better than their best member ($N=525$), {\Ctask}: constructive ($N=986$), (-): non-constructive ($N=464$), {\Wtask}: worse than worst member ($N=248$).
}
\end{figure*}

\subsection{Idea flow}

Task-oriented discussions are the the primary way of exchanging ideas and opinions between the group members; 
some are 
quickly
discarded while 
others
 prove useful to the final guess. The arrows in Figure~\ref{fig:example} show how
ideas are introduced and discussed
in that example conversation.
We attempt to capture the shape in which the ideas flow in the discussion.  In particular, we are interested in how many ideas are discussed, how widely they are adopted, 
 who tends to introduce them,
 and how.

We consider as candidate ideas all nouns, proper nouns, adjectives and verbs that are not 
 stopwords.  
 As soon as a candidate idea introduced by a player is {\em adopted} by another, we count it.
Henceforth, we'll refer to such adopted ideas simply as {\em ideas}.
In general chat domains, state-of-the-art 
 models of
conversation structure 
 use unsupervised probabilistic
models \cite{ritter2010unsupervised,elsner2010disentangling}. Since {\scr} conversations
are short and focused, the {\em adoption} filter is sufficient to accurately capture
what ideas are being discussed;
a manual examination of the ideas reveals almost exclusively place names
and 
 words such as {\em flag}, {\em sign}, {\em road}---highly relevant clues in the
context of {\scr}.
In Figure~\ref{fig:example}, three ideas are adopted: {\em China}, {\em buildings} and
{\em Shanghai}. The only idea adopted by all players is {\em buildings}, a good signal that
this was the most important clue.
A notable limitation is that this approach cannot capture the connections between {\em Shanghai} and {\em China}, or {\em buildings} and {\em apartments}.  Further work is needed to
robustly capture
such variations in idea flow, as they could reveal trajectories (discussion getting more specific or more vague) or lexical choice disagreement.

Balance in
idea contributions
between the team members is a good indicator of productive discussions.  In particular, in the best teams (the ones that outperform the best player, i.e., $c_{\text{best}}>0$) the
most idea-prolific player introduces fewer ideas, on average, than in the rest of the games (Figure~\ref{fig:maxideas},
$p=0.01$).\footnote{All p-values reported
reported in this section are based on one-sided Mann-Whitney rank statistical significance tests.}
In Figure~\ref{fig:example}, {\em E} is the most prolific player and only introduces two ideas.
To further capture the balance in contribution between the team members, we use the entropy of the number ideas introduced by each player.  We also count the number of ideas adopted unanimously as an indicator of convergence in the conversation.

In terms of the overall number of ideas discussed, both
the best teams (the ones that outperform the best player)
and the worst teams (the ones that perform worse than the worst player) discuss fewer ideas
than the rest (Figure~\ref{fig:nideas},
$p=0.006$).
Indeed, an ideal
interaction would avoid distracting ideas, but in 
teams with communication breakdowns, members might fail to adequately discuss the ideas that led them to
their individual guesses.

 The language used to introduce
new ideas can indicate confidence
or hesitation;
in Figure~\ref{fig:example},
a hedge ({\em would}) is used
when introducing the {\em buildings} cue.
We find that, in 
 teams that outperform the best player, 
 ideas are less likely to be  
accompanied by hedge words
when introduced
(Figure~\ref{fig:introhedge},
$p<10^{-4}$), 
showing less hesitation.
Furthermore, the level of confidence used when players adopt others' ideas is also informative (Figure~\ref{fig:adopthedge}).
Interestingly,
overall occurrences of certainty and hedge words (detailed in Section~\ref{sec:ling}) are not 
predictive, suggesting that ideas are good selectors for important discussion segments.
\subsection{Interaction dynamics}\label{sec:interdym}
\xhdr{Balance.} Interpersonal balance has been shown 
to be
predictive of team performance
\cite{jung2011engineering,Jung:2012:GHB:2207676.2208523} and, similarly, forms of linguistic balance have been
shown to characterize stable relationships \cite{niculae15betrayal}.
Here we focus on balance in contributions to the discussion and the decision process.
In search of measures 
applicable to 
teams of arbitrary sizes, we use binary indicators of whether all players participate in the discussion and 
in moving the marker,
  as well as whether
 at least two players move the marker.
To measure team balance with respect to continuous user-level features, we use the entropy of these features:
\[
\operatorname{balance}(S) = -\sum_{\bar{s} \in S} \bar{s} \log_{|S|}\bar{s},
\]
\noindent where, for a given feature, $S$ is the set of its values for each user,
normalized to
sum
to 1.  
For instance,
the chat message entropy
is 1 if everybody chats equally, and decreases toward 0 as one or more players dominate.
We use the entropy of
the
number of messages, words per message, and number of intermediate guesses.
In teams that outperform the best player,
users take turns controlling the marker more uniformly
(Figure~\ref{fig:entmove},
$p=0.006$),
adding further evidence that well-balanced teams perform best.

\xhdr{Language matching.} We investigate matching
at stopword, content word, and POS tag bigram level: the stopword
matching
at a turn is given by the number of stopwords from the earlier message repeated in the reply, divided by the total number of distinct stopwords to choose from; similarly for the rest.  We micro-average
over the
conversation:
\[ \operatorname{match} = \frac{\sum_{(\text{msg}, \text{reply}) \in \text{Turns}} |\text{msg} \cap \text{reply}|}{\sum_{(\text{msg}, \text{reply}) \in \text{Turns}} |\text{msg}|}.
\]
\noindent We also micro-average at the player-pair level, and use 
the maximum pair value
as a feature.  This gives an indication of how cohesive the closest pair
is, which can be
a sign of 
the level of
 power imbalance between the two \cite{danescu2012echoes}.
Figure~\ref{fig:maxaccom} shows that in
teams that outperform the best individual
the most cohesive pair
matches
fewer
content words
($p=0.023$).
Overall
matching
is also significant,
notably in terms of part-of-speech bigrams;
in teams that outperform
the best individual
there is less
overall
matching
(Figure~\ref{fig:meanaccom},
$p=0.007$).
These results suggest that 
in
constructive teams
the relationships
between the members
are less subordinate.

\xhdr{Agreement and confidence.} We capture the amount of agreement and disagreement using
high-precision keywords and filters validated on a subset of the data. (For instance, the word {\em sure} marks agreement if found
at the beginning of a message, but not otherwise.)
In Figure~\ref{fig:example}, agreement
signals
are underlined
with purple; the team exhibits no disagreement. 

The relative position of successive guesses 
made
 can 
also
 indicate whether the team is refining a guess or contradicting each other.
We measure the median distance between intermediate guesses,
as well as between guesses made by different players;
in constructive teams, the jumps between different player guesses are smaller
($p<10^{-16}$).
Before the discussion starts, players
are asked to self-evaluate their confidence
in their individual 
guesses.
Constructive teams 
have more
 confident members on average
($p<10^{-5}$).

\subsection{Other linguistic features}
\label{sec:ling}
\xhdr{Length and variation.}  We measure the average number of words per message, the total number of words used to express the {\em solo phase} reasons, and the overall type/token ratio of the conversation.  We also measure responsiveness in terms of 
the mean time between turns and the total number of turns.

\xhdr{Psycholinguistic lexicons.}  We use hand-crafted lexicons inspired from LIWC
\cite{tausczik2010psychological} to capture certainty and pronoun use. For example,
the conversation in Figure~\ref{fig:example} has two confident phrases, underlined in red.
We also use a custom hedging lexicon adapted from \newcite{hyland2005metadiscourse}
for conversational data; hedging words are underlined in blue in Figure~\ref{fig:example}. 
To estimate how grounded the conversation is, we
measure the average {\em concreteness} of all content nouns, adjectives, adverbs and
verbs, using scalar word and bigram ratings from \newcite{brysbaert2014concreteness}.%
\footnote{We scale the ratings to lie in $[0, 1]$. 
We extrapolate to out-of-vocabulary words
by regressing on dependency-based word embeddings
\cite{levy2014dependencybased};
this approach is highly accurate
(median
absolute error of about 0.1).}
Concreteness reflects the degree to which a word denotes something perceptible, as opposed to
ideas and concepts. 
Words like
{\em soil} and {\em coconut} are highly concrete,
while
words like
{\em trust}
have low concreteness.

\xhdr{Game-specific words.} We put together a lexicon of geography terms and place names, to capture task-specific
discussion. We use a small set of
words specific to the {\scr} interface, such as
{\em map, marker,} and {\em game}, to capture 
phatic
 communication.
Figure~\ref{fig:geolex} shows that constructive teams tend to use more geography terms
($p=0.008$), possibly because of more on-topic discussion and a more focused vocabulary.

%\xhdr{POS, BOW.} 
\xhdr{Part-of-speech patterns.}  We use n-grams of coarse part-of-speech tags as a general way
of capturing common syntactic patterns.  

\begin{table}
\small
\centering
\begin{tabular}{l l l l l l l}
\toprule
  & \multicolumn{3}{c}{Full conversation} & \multicolumn{3}{c}{First 20s} \\
Features & {\Btask} & {\Ctask} & {\Wtask} & {\Btask} & {\Ctask} & {\Wtask} \\
\midrule
Baseline 	 & .51 & .52 & .55 & .52 & .50 & .54 \\
\sep
  Linguistic & .54 & .52 & .50 & .50 & .51 & .50 \\
 Interaction & .55$^\dagger$ & .56$^\dagger$ & .53 & .55$^\dagger$ & .57$^\star$ & .56 \\
         POS & .55$^\dagger$ & .59$^\star$ & .55 & .54 & .54 & .53 \\
\sep
    {\bf All} & .56$^\star$ & .60$^\star$ & .56 & .56$^\star$ & .57$^\star$ & .57$^\dagger$ \\
\bottomrule
\end{tabular}
\caption{\label{tab:results}
Cross-validation AUC scores.
Significantly better than chance scores after 5000 
permutations denoted with $\star$~($p<0.05$) and $\dagger$~($p<0.1$).
}
\end{table}

\section{Predicting constructiveness}\label{sec:predict}
\subsection{Experimental setup}
So far we have characterized the relation between a team's interaction patterns and its level of productivity.
This opens the
door
towards recognizing constructive and non-constructive interactions in realistic settings where
the true answer is not known.  Ideally, such an automatic system could prompt unproductive {\teams} to reconsider their approach, or to aggregate their
individual answers instead.
With {\em early detection},
non-constructive discussions
could
be stopped
or steered on
the right
track.
In order to assess the feasibility of such a challenging task and to compare the predictive power of our features,
we consider three classification objectives: \\
{\small
{\Btask}: Team outperforms its best member ($c_{\text{best}}>0$)?\\
{\Ctask}: Team is constructive ($c_\text{avg}>0$)?\\
{\Wtask}: Team underperforms its worst member ($c_\text{worst}<0$)?}
\noindent To investigate
{\em early detection}, we evaluate the
classification performance when using data from only the first 20 seconds of the
team's
interaction.\footnote{%
Measured
from the first chat message or
guess.
For this evaluation, we remove teams 
where
the first 20 seconds contain over
75\% of the interaction,
to avoid
distorting the results with teams who
make their
decision early, but take longer to submit.
The 20 second threshold was chosen as a trade-off in terms of how much interaction it covers in the games.
}

Since all three objectives are imbalanced (Figure~\ref{fig:constr}), we use the area
under the ROC curve (AUC)
as the performance
metric, and we use logistic regression models.
We perform 20 iterations of
{\puzzle}-aware
shuffled train-validation splitting,
followed by 5000 iterations on the best models, to estimate variance.
This ensures that the models don't learn to overfit
{\puzzle}-specific
signals.
The combined model uses weighted model averaging.  We 
use
grid search for regularization
parameters, feature extraction parameters,
and
combination weights.
\subsection{Discussion of the results (Table~\ref{tab:results})}
We compare to a baseline consisting of the team size, average number of messages per player,
and conversation duration.
For comparison, a bag-of-words classifier does no better than chance and is on
par with the baseline.
We refer to idea flow and interaction dynamics features (Section~\ref{sec:interdym}) as {\em Interaction}, and to linguistic and lexical features (Section~\ref{sec:ling}) as {\em Linguistic}.
The combination model including baseline,
interaction,
linguistic
and
part-of-speech n-gram features, is consistently the best and significantly outperforms random guessing (AUC .50)
in nearly all settings.
While overall scores are 
modest,
 the results confirm that our conversational analysis framework
has predictive power, and that the high-stakes task of {\em early prediction} is 
feasible.
The language
used when introducing and adopting ideas, together with balance and language matching features, are selected
in nearly all settings.
The least represented class {\Wtask} has the highest variance in prediction, suggesting that more data collection is needed to successfully capture extreme cases.
Useful POS patterns capture the amount of proper nouns and 
their
contexts: proper nouns at the end of messages
are indicative of constructiveness, while proper nouns followed by verbs are a negative feature.
(The constructive discussion shown in Figure~\ref{fig:example} has most proper nouns at the end of messages.)

A manual error analysis of the
false positives and false negatives where our best model is most confident points
to games with very short conversations and spelling mistakes, confirming that the noisy data problem causes learning
and modeling
difficulties.

\section{Conclusions and Future Work}\label{ref:conclusion}

We developed a framework based on conversational dynamics 
in order to distinguish between productive and unproductive task-oriented discussions.
By applying it to an online collaborative game we designed for this study, we reveal new interactions with conversational patterns. 
Constructive teams are generally well-balanced on multiple aspects, with teammembers participating equally in proposing ideas and making guesses and showing little asymmetry in language matching.
Also,
the flow of ideas
 between 
teammates
marks predictive linguistic cues,
with
the most constructive teams using fewer hedges when introducing 
and adopting
ideas.
We show that such cues
have predictive power even
when extracted from
the first 20 seconds of the conversations.
In future work, improved classifiers could lead to 
a system that can
intervene in non-constructive discussions early on, steering them on track
and preventing wasted time. 

Further improving classification performance on such a difficult task
will hinge on better conversation processing tools, adequate for the domain and robust to the informal language style.  In particular, we plan to develop and evaluate models for idea flow and (dis)agreement, using more advanced features (e.g., from dependency relations and knowledge graphs).

The {\scr} game is continuously accumulating more data, enabling further
development on conversation analysis.  Our full control over the game permits
manipulation and intervention experiments that can further advance research
on teamwork. 
 In future work, we envision
 applying our framework to settings where
teamwork
 takes place online, such as open-source software development, Wikipedia editing, or massive open online courses.

\vspace{0.05in}
\xhdr{Acknowledgements} We are particularly grateful to Bob West for the
poolside
chat that inspired the design of the game, to Daniel Garay, Jinjing Liang and Neil Parker for participating in its
development, and to the numerous passionate players.
We are also grateful to
Natalya Bazarova,
David Byrne,
Mala Gaonkar,
Lillian Lee,
Sendhil Mullainathan,
Andreas Veit, %
Connie Yuan,
Justine Zhang
and the anonymous reviewers for their insightful
suggestions.  This work was supported in part by a Google Faculty Research Award.

\bibliography{construct_clean}
\bibliographystyle{naaclhlt2016}

\end{document}